%% file: main.tex
\documentclass[10pt,twocolumn,letterpaper]{article}

\usepackage{cvpr}              

\input{preamble}

\definecolor{cvprblue}{rgb}{0.21,0.49,0.74}
\usepackage[pagebackref,breaklinks,colorlinks,allcolors=cvprblue]{hyperref}

\def\paperID{6636} 
\def\confName{CVPR}
\def\confYear{2026}

\title{Locate‑Then‑Examine: Grounded Region Reasoning Improves Detection of AI‑Generated Images}

\author{Yikun Ji$^{1,2}$ \quad Yan Hong$^2$ \quad Bowen Deng$^1$ \quad Jun Lan$^{2}$\footnotemark[1] \quad Huijia Zhu$^2$ \quad Weiqiang Wang$^2$ \\ Liqing Zhang$^{1}$\footnotemark[1] \quad Jianfu Zhang$^{1}$\footnotemark[1]\\
$^1$Shanghai Jiao Tong University \qquad $^2$Ant Group \\
{\tt\small \{da-kun, lqzhang, c.sis\}@sjtu.edu.cn \quad yelan.lj@antgroup.com}
}

\begin{document}
\maketitle

\begingroup
\renewcommand{\thefootnote}{\fnsymbol{footnote}}
\footnotetext[1]{Corresponding authors.}
\endgroup

\input{sec/0_abstract}
\input{sec/1_intro}
\input{sec/2_related}
\input{sec/3_method}
\input{sec/4_exper}
\input{sec/5_conclusion}

\section{Acknowledgments}
This work was supported in part by the National Natural Science Foundation of China (Grant Nos. 62302295, 62595733, and 62561160155), the Shanghai Municipal Science and Technology Major Project (Grant No. 2021SHZDZX0102). This work was also supported by Ant Group.

{
    \small
    \bibliographystyle{ieeenat_fullname}
    \bibliography{main}
}

\clearpage
\input{sec/X_suppl}
{
    \small
    \IfFileExists{main.bbl}{\input{main.bbl}}{}
}

\end{document}

%% file: preamble.tex
\newcommand{\TODO}[1]{\textbf{\color{red}[TODO: #1]}}
\renewcommand{\TODO}[1]{}

\usepackage{microtype}

\input{math_commands}

\usepackage{graphicx}
\usepackage{amssymb}
\usepackage{amsmath}
\usepackage{wrapfig}
\usepackage{booktabs}
\usepackage{multirow}
\usepackage[table,xcdraw]{xcolor}
\usepackage[normalem]{ulem}
\useunder{\uline}{\ul}{}
\usepackage[bbgreekl]{mathbbol}
\DeclareSymbolFontAlphabet{\mathbbl}{bbold}
\usepackage{caption}
\usepackage{enumitem}

%% file: math_commands.tex
\usepackage{amsmath,amsfonts,bm}

\def\eqref#1{equation~\ref{#1}}

\def\1{\bm{1}}

\DeclareMathAlphabet{\mathsfit}{\encodingdefault}{\sfdefault}{m}{sl}
\SetMathAlphabet{\mathsfit}{bold}{\encodingdefault}{\sfdefault}{bx}{n}



%% file: sec/0_abstract.tex
\begin{abstract}
The rapid growth of AI-generated imagery has blurred the boundary between real and synthetic content, raising practical concerns for digital integrity. Vision-language models (VLMs) can provide natural language explanations, but standard one-pass classifiers often miss subtle artifacts in high-quality synthetic images and offer limited grounding in the pixels. We propose \textbf{Locate-Then-Examine} (LTE), a two-stage VLM-based forensic framework that first localizes suspicious regions and then re-examines these crops together with the full image to refine the real vs. AI-generated verdict and its explanation. LTE explicitly links each decision to localized visual evidence through region proposals and region-aware reasoning.
To support training and evaluation, we introduce \textbf{TRACE}, a dataset of 20,000 real and high-quality synthetic images with region-level annotations and automatically generated forensic explanations, constructed by a VLM-based pipeline with additional consistency checks and quality control. 
Across TRACE and multiple external benchmarks, LTE achieves competitive accuracy and improved robustness while providing human-understandable, region-grounded explanations suitable for forensic deployment.
\end{abstract}

%% file: sec/1_intro.tex
\section{Introduction}
\label{sec:intro}

Recent advances in image generation~\citep{wang2025designdiffusion,li2025fair,chadebec2025flash} have pushed photorealism to a point where synthetic content is difficult to distinguish from real imagery. As these models are increasingly adopted in consumer and creative platforms, the need for reliable and interpretable forensic tools has become more urgent.
Existing detection systems are predominantly classification-based: they can achieve high accuracy on curated datasets, but their decision process is opaque and their robustness is limited.
Models trained to detect generated images from one architecture often underperform on unseen generators.

\begin{figure*}[t]
    \centering
    \includegraphics[width=0.88\linewidth]{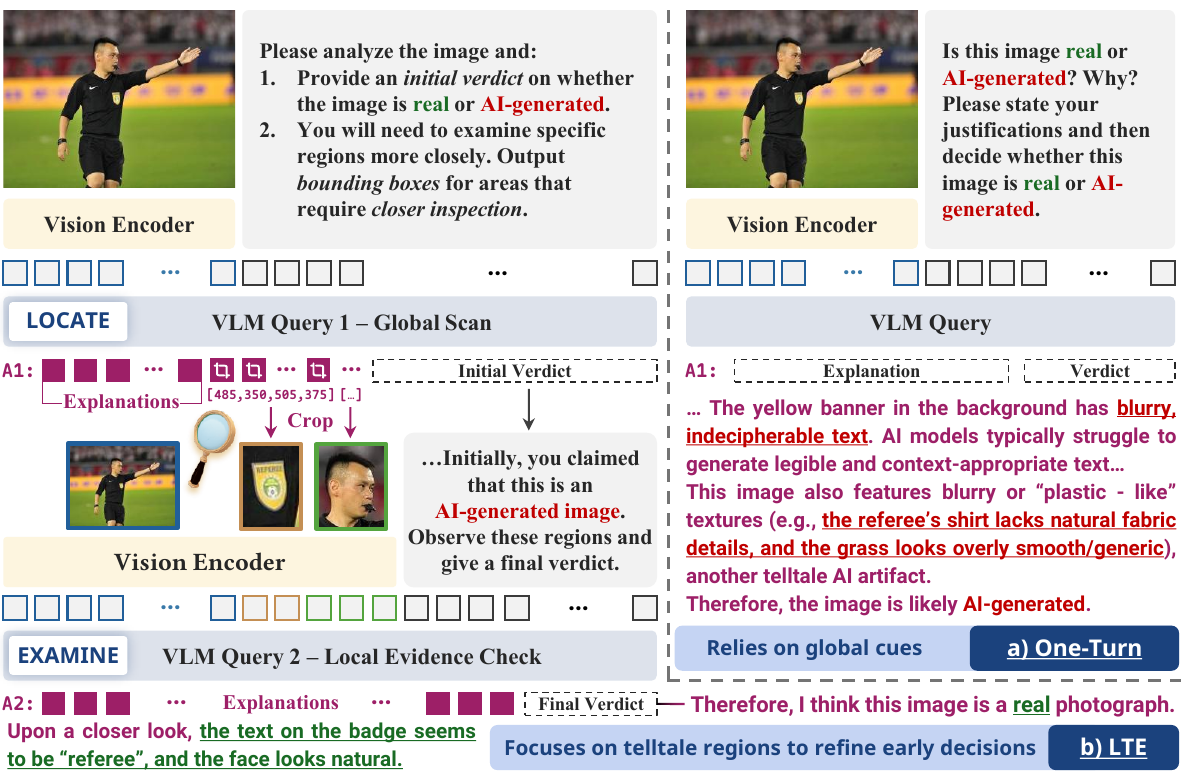}
    \caption{
    (a) Without revisiting specific details, VLMs may overlook critical cues and produce false reasoning with incorrect decisions.
    (b) Our two-stage \textbf{LTE} pipeline. The VLM first performs a global scan to query region(s) of interest (Query~1), then analyzes the cropped regions for a detailed, final verdict with grounded explanations (Query~2, ``Local Evidence Check'').
    }
    \vspace{-8pt}
    \label{fig:zoomin}
\end{figure*}

The rise of Vision-Language Models (VLMs)~\citep{fang2025guided,chen2024internvl,man2025argus,wu2025combating,wu2025grounded,zhang2025critic} has opened a new frontier, offering a promising path towards semantic-level analysis and greater interpretability. Recent approaches re-frame the detection task as a visual question answering problem~\citep{FakeReasoning} or an image captioning task~\citep{Bi-LORA}. 
However, to reach high accuracy, these methods typically rely on external modules such as segmentation or classification heads~\citep{AIGIHolmes,LEGION}, which underutilize the intrinsic knowledge and common-sense reasoning embedded in the VLMs and reduce them to passive feature extractors. 
More fundamentally, most approaches perform a single global pass over the image: the visual encoder compresses the scene into a limited set of tokens, attention is spread across the entire image, and fine-grained forensic cues (\textit{e.g.}, tiny text artifacts, stitching seams, periodic textures, specular edges) are weakened by downsampling and pooling~\citep{ComFor, LLaVA}. 
Without a mechanism to revisit localized regions and verify hypotheses, these subtle but decisive artifacts are easily missed, leading to unstable decisions on high-quality synthetic images.
As illustrated in Fig.~\ref{fig:zoomin}(a), insufficient localization can cause the model to rely on prior expectations rather than pixel-level evidence, resulting in unreliable reasoning and incorrect outcomes.

In this work, we introduce a region-grounded reasoning framework, \emph{Locate-Then-Examine (LTE)}, that moves beyond single-pass classification. Instead of relying on a global prediction, our approach equips a VLM with the ability to \textit{localize} suspicious regions and then \textit{re-examine} these regions together with the full image to refine its decision. This design integrates the VLM’s semantic reasoning with targeted visual inspection, enabling the system to generate hypotheses about potentially synthetic cues and verify them at higher resolution.
To support this process, we construct a dataset, \textit{Text‑Region Aligned Cues for Explainable AIGI detection (TRACE)}, that provides region crops coupled with grounded forensic explanations for supervised training. As shown in Fig.~\ref{fig:zoomin}(b), this targeted locate-then‑examine mechanism produces more reliable evidence attribution and more stable decisions on challenging synthetic images.
Our experiments demonstrate that this iterative, region-aware analysis improves both robustness and interpretability across diverse generative models and external benchmarks.
Our major contributions are threefold:
\begin{enumerate}[leftmargin=*]
\item{\textbf{Region-grounded two-stage forensic reasoning.}} We propose \textit{LTE}, a VLM-based framework that first identifies suspicious regions and then re-examines these regions to produce a refined, evidence-grounded verdict.
\item{\textbf{An evidence-aligned forensic dataset.}} We build \textit{TRACE}, a 20,000-image dataset with region-level annotations and forensic explanations, generated through a VLM-based pipeline with consistency checks and quality control.
\item{\textbf{Robust and interpretable detection.}} LTE improves accuracy and stability across TRACE and multiple external benchmarks while providing region-grounded explanations, including reduced misclassification rates by 38.2\% for the 7B variant and 67.4\% for the 32B variant.
\end{enumerate}

%% file: sec/2_related.tex
\section{Related Works}

\noindent\textbf{Detection of AI-generated images.}
Image forgery detection has evolved from traditional feature engineering to modern deep learning methods. Early detection methods rely on handcrafted features and statistical anomalies~\citep{TraditionalForgery1,TraditionalForgery2,TraditionalForgery3} or perform frequency-domain analysis for identifying GAN-specific artifacts~\citep{FrePGAN}. The advent of deep learning marked a paradigm shift, with CNN-based detectors such as CNNSpot~\citep{CNNSpot} demonstrating remarkable generalization capabilities across various GAN architectures when trained on ProGAN-generated~\citep{ProGAN} images. As generative models evolved beyond GANs to include diffusion models~\citep{DDIM,DDPM,le2025one,ye2025schedule}, detection strategies adapted accordingly~\citep{DMImageDetection, NPR, ComFor, AEROBLADE}. Notably, DIRE~\citep{DIRE} pioneered the use of reconstruction error metrics specifically tailored for diffusion-generated content. NPR~\citep{NPR} leveraged frozen CLIP encoders to maintain domain invariance.

Despite these advances, explainability and robust generalization remain major challenges. The emergence of VLMs offers a new frontier by enabling semantic-level analysis and natural language reasoning.
Many approaches re-formulate this classification problem to VQA questions~\citep{AntifakePrompt, ABench} or image captioning tasks~\citep{Bi-LORA}. Several forensics datasets~\citep{FakeBench, ABench, FakeReasoning} are curated using VLMs, creating training corpora that combine visual analysis with natural language reasoning.
Specifically, ~\citet{AIGIHolmes} combines NPR~\citep{NPR} with LLM, achieving high detection accuracy with good interpretability.
In terms of localization, FakeShield~\citep{FakeShield} introduces the Segment Anything~\citep{SAM} module to acquire the tampered mask for the manipulated image.
LEGION~\citep{LEGION} postfixes the vision encoder with an MLP to identify the authenticity of the input image.
Our work builds on this direction by introducing spatial grounding and iterative refinement, transforming VLMs from passive analyzers into active visual investigators that fully leverage their inherent common-sense reasoning.

\noindent\textbf{Training and Fine-Tuning Reasoning-Capable VLMs.}
Improving the reasoning abilities of VLMs is essential for tasks demanding sophisticated comprehension~\citep{wu2025icm,wu2025combating,yang2025heie}. Early approaches focused on transforming images into structured textual representations to facilitate language-based reasoning~\citep{R1-OneVision}. More recent studies have emphasized cultivating advanced cognitive skills, such as self-verification, self-correction, fostering ``slow thinking'' capabilities~\citep{VL-Rethinker}, and regulating reasoning depth to mitigate issues like ``overthinking''~\citep{Fast-Slow-Thinking}. Additionally, efforts have concentrated on developing high-quality multi-modal Chain-of-Thought (CoT) datasets~\citep{Vision-R1} to steer the reasoning process effectively.
DeepSeek-Math~\citep{DeepSeekMath} provides a solid foundation and methodology for fine-tuning large language models.
For reward design, in addition to the outcome reward used in DeepSeek-Math, expert LLMs are also widely used as an online training reward provider~\citep{RewardBench}.
Researchers are also experimenting with using IoU~\citep{MedVLM-R1} and BLEU~\citep{BLEUBERI} metrics as rewards.
Building on these advances, we train our VLM with spatially grounded supervision and iterative reasoning objectives, where cropped regions and fine-grained explanations guide the model to link decisions with explicit visual evidence, thereby enhancing both accuracy and interpretability.

%% file: sec/3_method.tex
\begin{figure*}[t]
    \centering
    \includegraphics[width=0.98\linewidth]{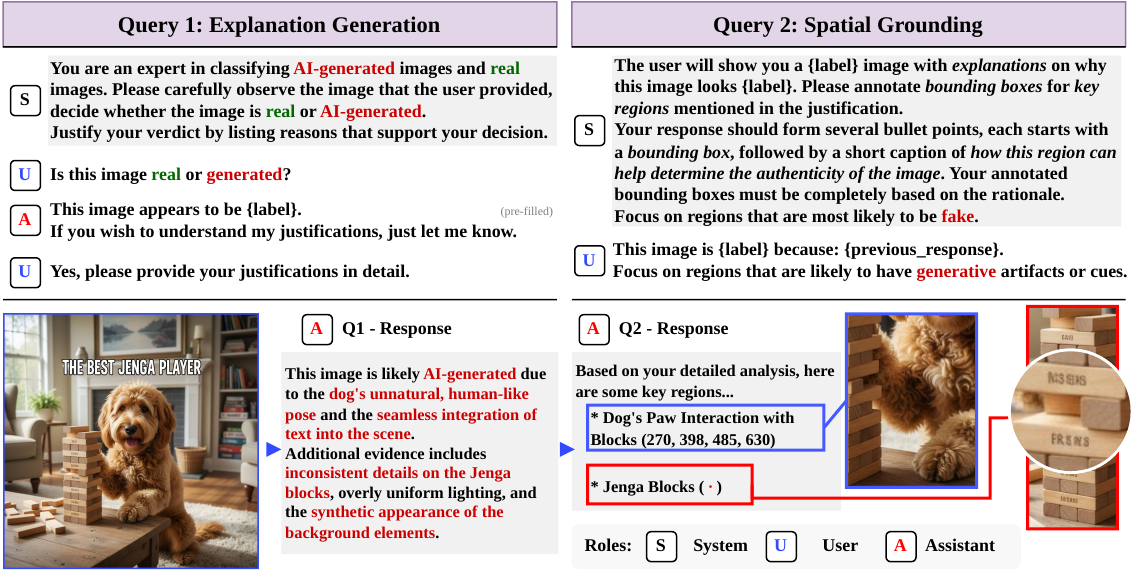}
    \caption{The proposed data annotation pipeline. We ask the forensics expert VLM in Query 1 ``Explanation Generation'' to identify key reasons that make this image look real or AI-generated, followed by Query 2 ``Spatial Grounding'', which uses the explanation to extract bounding boxes.}
    \label{fig:dataset}
    \vspace{-8pt}
\end{figure*}

\section{Methodology}
\subsection{Locate, Examine and Verdict}

We argue that single-pass analysis is often insufficient for reliable forensics: it tends to overlook subtle flaws and produce generic, non-specific explanations (e.g., ``global saturation,'' ``blurry textures,'' etc.) that are weakly tied to decisive artifacts.
In practice, the most informative forensic cues are frequently confined to small regions, and both VLMs and humans require focused, high-resolution inspection to detect them.
Our two-stage approach directly addresses this: (1) a \emph{Locate with Global Scan} stage inspects the entire image to hypothesize and localize potentially decisive regions; (2) an \textit{Examine with Local Evidence Check} stage re-examines these candidate regions together with the full image, performing fine-grained analysis to extract explicit visual evidence and produce the final verdict.
We hope that when the global view of an image is uncertain, the model revisits and focuses on regions that are most likely to contain decisive forensic cues.
Figure~\ref{fig:zoomin} shows the two-stage inference workflow. The setup is detailed below:

\noindent\textbf{Query 1: Locate with Global Scan.}
Given an input image $I$, we prompt a grounding-capable VLM to perform comprehensive visual analysis. The model generates, in order:
\begin{itemize}[leftmargin=*]
    \item Preliminary explanation $E_1$ articulating the reasoning.
    \item Suspicious regions $B = \{b_1, ..., b_n\}$ where $b_i = (x_1, y_1, x_2, y_2)$ represents bounding box.
    \item Initial verdict $v_1 \in \{\text{real}, \text{generated}\}$.
\end{itemize}
This stage leverages the VLM’s ability to process global context while highlighting locally anomalous regions.
However, limited resolution and tokenization can still weaken subtle artifacts, which motivates a second, magnified pass over the most informative areas.
Recent work on tool-augmented VLMs shows that equipping models with visual operations and step-wise interactions over images enables more structured reasoning and better interpretability~\citep{ThinkingWithImages,shen2025satori,fan2025grit,su2025openthinkimg,cheng2025visual}.
Our framework instantiates this idea for AI-image forensics.

Figs.~\ref{fig:zoomin} and \ref{fig:dataset} illustrate the kinds of regions that warrant closer inspection for real and fake cases.
We focus on two main types: (i) \textit{regions inherently challenging for generative models}, such as human faces or hands (Fig.~\ref{fig:zoomin}), and fine-grained animal attributes like paws or poses (Fig.~\ref{fig:dataset}); and (ii) \textit{image-specific details that are difficult to reproduce}, including logos on a referee’s shirt (Fig.~\ref{fig:zoomin}) or small texts (Fig.~\ref{fig:dataset}).
By first locating and then examining these regions at higher resolution, the model reduces global uncertainty, improves prediction accuracy, and yields more reliable explanations.

\noindent\textbf{Query 2: Examine with Local Evidence Check.}
For each identified region $b_i$, we first extract crops $C_i = \text{Crop}(I, b_i)$, and then provide the VLM with both the original image $I$ and the crop collection $\{C_i\}_{i=1}^n$, enabling comparative analysis between global context and local details. This dual-input mechanism produces, in order:
\begin{itemize}[leftmargin=*]
    \item Refined explanation $E_2$ grounded in specific visual evidence.
    \item Final verdict $v_2 \in \{\text{real}, \text{generated}\}$.
\end{itemize}
By providing these image crops $C_i$, we enrich the input with fine-grained visual tokens, allowing the model to correct earlier misjudgments through magnified analysis, much like a forensic expert using a ``magnifying glass''. This second-stage verification substantially improves both accuracy and explanation quality.
Supporting such region-aware reasoning requires training data with fine-grained annotations, which motivates the construction of our \textbf{\textit{TRACE}} dataset and the training of \textbf{\textit{LTE}}.

\subsection{TRACE Dataset Construction}

Our two-stage pipeline requires localization information for both real and AI-generated images.
As discussed earlier, such information should highlight (i) regions inherently challenging for generative models and (ii) image-specific regions that are difficult to reproduce.
While the locate-then-examine process mimics the intuition of human forensic experts, VLMs do not naturally possess this capability. To enable this behavior and improve both grounding accuracy and detection performance, we design a targeted training strategy based on $(I, y, E, B)$ tuples, where $y\in\{\text{real},\text{generated}\}$ is the ground-truth label, providing supervision for classification and grounding tasks within the pipeline. Image crops $C$ are deterministically derived from $(I, B)$.
To acquire grounding-aware training data, we create the \textbf{\textit{TRACE}} dataset by leveraging different VLM experts to generate textual explanations and bounding boxes.
Previous works~\citep{FakeReasoning, ExplainableMLLMDefake} and our preliminary tests have confirmed that the OpenAI GPT-4o can produce detailed forensic explanations describing why an image is real or AI-generated.
Additionally, the Qwen-2.5-VL series of VLMs~\citep{Qwen25VL} can extract spatial regions from these explanations and output bounding boxes, making them well-suited for generating spatially grounded annotations.
An overview of the pipeline is provided in Figure~\ref{fig:dataset}, showing the prompts used in the automated annotation process. 
We address the lack of spatially grounded forensic annotations through an automated pipeline as follows:
\begin{enumerate}[leftmargin=*]
\item{\textbf{Explanation Generation}.} For images with known labels, GPT-4o generates forensic explanations focusing on specific visual evidence.
The prompt used at this stage is designed to elicit detailed reasoning about the depicted objects, arrangement, perspective, and other relevant aspects of the given images.
\item{\textbf{Spatial Grounding.}} Qwen-2.5-VL extracts bboxes from the explanations, creating $(I, y, E, B)$ tuples.
\end{enumerate}

\noindent\textbf{Data Purification.}
While Qwen-2.5-VL is generally effective at grounding forensic cues, it occasionally produces bounding boxes that are suboptimal for training.
In particular, the model may generate boxes covering more than 50\% of the image, often corresponding to global image imperfections rather than localized artifacts.
In other cases, the model may regress to object detection, returning boxes that fully encapsulate the primary object when only a small portion contains the relevant flaw.
To ensure that TRACE contains high-quality and region-specific annotations, we apply a multi-stage purification process.
First, we perform explanation–region consistency checks.
For each image, GPT-4o generates two independent explanations, and semantic similarity is computed between them. Explanations with low agreement are discarded.
Likewise, Qwen-2.5-VL is prompted twice for spatial grounding, and bounding boxes are kept only when the overlap between runs exceeds an IoU threshold.
Samples in which the explanation refers to a specific cue that is not spatially covered by any bounding box are removed.
This cross-VLM validation mitigates bias from any single model and filters inconsistent $(E, B)$ pairs.
Second, we remove bounding boxes that are excessively large or misaligned.
For each region $b_i$, we evaluate whether its area exceeds 50\% of the image or whether it fully encloses the main object, indicating a fallback to object detection.
Such boxes typically do not correspond to fine-grained forensic evidence and are therefore discarded.

\begin{figure*}[t]
    \centering
    \includegraphics[width=1\linewidth]{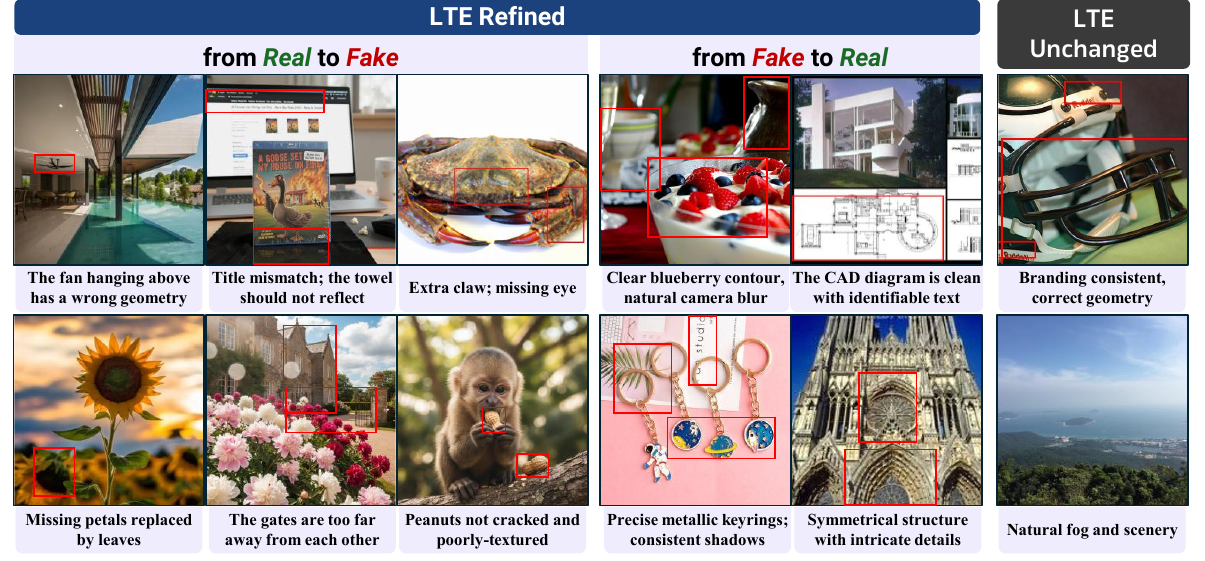}
    \caption{Examples from the test set of TRACE, captions are summarized from the Query 2 response, generated by LTE-32B.}
    \label{fig:samples}
     \vspace{-16pt}
\end{figure*}

\noindent\textbf{Image Distribution.} TRACE consists of 10,000 real images and 10,000 AI-generated images. All of which are annotated with explanations and spatial grounding, among which 99.5\% of the images have at least one bounding box, with an average of 3.24 bounding boxes per image after filtering.
The real images are sourced equally from ImageNet~\citep{ImageNet} and COCO~\citep{COCO}, and the AI-generated images are equally sourced from GPT-Image-1~\citep{GPT-Image-1} and Gemini 2.5 Flash Image~\citep{NanoBanana}.

\subsection{Training Procedure of LTE}

With $(I, y, E, B)$ tuples from \textbf{\textit{TRACE}}, we can fine-tune a VLM to become an expert in the two queries.
Inspired by DeepSeek-Math~\citep{DeepSeekMath}, our fine-tuning approach employs a two-phase training paradigm that combines supervised fine-tuning (SFT) with reinforcement learning (RL) implemented through Group Relative Policy Optimization (GRPO).

\noindent\textbf{Supervised Fine-tuning Phase.}
The training begins with SFT to establish foundational capabilities and ensure stable model behavior. During this phase, all trainable parameters across the model's vision encoder, projection layers, and language modeling components undergo optimization using supervised signals from the dataset.
This initial phase serves two critical purposes: (1) teaching the model to generate outputs that conform to our specified structured format, and (2) establishing baseline performance before applying reinforcement learning techniques.

\begin{table*}[t]
\centering
\caption{Detection accuracy and reasoning quality metrics on the TRACE test set. 
E- stands for explanation-only, and E+G- stands for explanation and grounding-only.}
\label{tab:result-combined}
\resizebox{0.8\linewidth}{!}{%
\begin{tabular}{@{}lcccccccc@{}}
\toprule
Method & Acc. & I-Acc. & C-Acc. & C-Cases (\%) & BLEU-1 & BLEU-2 & ROUGE-L & IoU \\ \midrule
\textbf{LTE-7B} & 0.942 & 0.866 & 0.915 & 9.2 & 0.314 & 0.209 & 0.291 & 0.316 \\
\textbf{LTE-32B} & \textbf{0.972} & 0.873 & 0.956 & 10.9 & 0.346 & 0.211 & 0.327 & 0.359 \\ \midrule
E-7B (one-turn) & 0.855 & - & - & - & 0.280 & 0.146 & 0.301 & - \\
E-32B (one-turn) & 0.869 & - & - & - & 0.294 & 0.153 & 0.315 & - \\
E+G-7B (one-turn) & 0.906 & - & - & - & 0.275 & 0.136 & 0.269 & 0.245 \\
E+G-32B (one-turn) & 0.914 & - & - & - & 0.282 & 0.149 & 0.295 & 0.254 \\ \midrule
Base-7B & 0.553 & - & - & - & 0.110 & 0.032 & 0.073 & - \\
Base-32B & 0.587 & - & - & - & 0.102 & 0.043 & 0.079 & - \\
SFT-7B & 0.719 & 0.724 & 0.447 & 4.8 & 0.156 & 0.042 & 0.129 & 0.094 \\
SFT-32B & 0.715 & 0.706 & 0.584 & 5.3 & 0.159 & 0.076 & 0.130 & 0.105 \\\midrule
No BLEU Reward-32B & 0.929 & 0.868 & 0.871 & 8.2 & 0.187 & 0.095 & 0.153 & 0.296 \\
Verdict Only-32B (one-turn) & 0.897 & 0.897 & - & - & - & - & - & - \\
Dual Verdict Reward-32B & 0.944 & 0.948 & 0.473 & 7.4 & 0.276 & 0.164 & 0.252 & 0.260 \\
Random Cropping-32B & 0.842 & 0.873 & 0.421 & 19.6 & 0.105 & 0.043 & 0.109 & - \\
Largest 4 Bboxes-32B & 0.934 & 0.873 & 0.938 & 7.0 & 0.303 & 0.182 & 0.158 & - \\
Largest 3 Bboxes-32B & 0.924 & 0.873 & 0.919 & 6.1 & 0.299 & 0.183 & 0.147 & - \\ \midrule
FakeShield~\citep{FakeShield} & 0.801 & - & - & - & 0.097 & 0.056 & 0.067 & 0.096 \\
LEGION~\citep{LEGION} & 0.654 & - & - & - & 0.102 & 0.058 & 0.054 & 0.061 \\ \bottomrule
\end{tabular}%
}
\vspace{-16pt}
\end{table*}

\noindent\textbf{Reinforcement Learning with Enhanced Rewards.}
Following SFT, we implement RL through two GRPO stages.
Our reward design extends beyond traditional classification and localization metrics by incorporating linguistic quality assessment through BLEU scores~\citep{BLEUBERI}.
We define reward functions tailored to two distinct query stages, each addressing specific aspects of the model's performance.
In the first query, the model generates an initial hypothesis, focusing on format compliance and localization precision. The reward structure is defined as follows:
\begin{itemize}[leftmargin=*]
    \item \textbf{Format Compliance:} The reward for correct output formatting is given by:
    \begin{small}\begin{equation}
        \mathcal{R}_F = 1 \text{ if output contains valid } \verb|<verdict>| \text{ tags}.
    \end{equation}\end{small}
    \item \textbf{Localization Precision:} Grounding accuracy is measured by IoU, rewarding precise spatial alignment. With $b_i$ denoting model-predicted boxes and $\hat{b}_j$ annotated boxes, we compute:  
    \begin{small}\begin{equation}
        \mathcal{R}_{\text{IoU}} = \frac{1}{|B|}\sum_{i} \max_j \operatorname{IoU}(b_i, \hat{b}_j).
        \label{eq:iou}
    \end{equation}\end{small}
\end{itemize}
This reward structure ensures the model prioritizes accurate spatial localization and adherence to the expected output format during the hypothesis generation phase.

In the second query, the model refines its hypothesis, emphasizing correct verdict prediction and high-quality explanation generation. The reward structure comprises:
\begin{itemize}[leftmargin=*]
    \item \textbf{Classification Accuracy:} The binary reward for correct verdict prediction is defined as:
    \begin{small}\begin{equation}
        \mathcal{R}_C = \mathbbl{1}[v_2 = y].
    \end{equation}\end{small}
    \item \textbf{Explanation Quality:} To encourage contextually appropriate explanations, we compute BLEU scores between generated explanations and reference texts:
    \begin{small}\begin{equation}
        \mathcal{R}_{\text{BLEU}} = \operatorname{BLEU}_2(E', E_{\text{ref}}).
    \end{equation}\end{small}
    where $E'$ is the explanation text from the model's output, and $E_{\text{ref}}$ represents the reference explanation from our annotated dataset.
\end{itemize}
This reward structure drives the model to produce accurate verdicts and coherent, high-quality explanations during the refinement stage.
Section~\ref{sec:ablation} shows the model performance when BLEU reward is ablated, or when a correct verdict counts as a reward in the Locate query.

%% file: sec/4_exper.tex
\section{Experiments}

\subsection{Setup}

\begin{table}[t]
\centering
\caption{Accuracy (\%) of LTE and other comparing methods on OoD datasets.}
\label{tab:result-ood}
\resizebox{1.0\linewidth}{!}{%
\begin{tabular}{@{}ccccccc@{}}
\toprule
\multirow{2}{*}{Datasets} & \multicolumn{2}{c}{MMFR} & \multicolumn{2}{c}{SynthScars} & \multicolumn{2}{c}{FakeClue} \\ \cmidrule(l){2-7} 
 & Acc. & BLEU-2 & Acc. & BLEU-2 & Acc. & BLEU-2 \\ \midrule
\textbf{LTE-32B} & \textbf{0.893} & \textbf{0.155} & 0.852 & 0.130 & \textbf{0.903} & \textbf{0.172} \\
\textbf{LTE-7B} & 0.892 & 0.151 & 0.826 & 0.135 & 0.871 & 0.157 \\ \midrule
LEGION~\citep{LEGION} & 0.193 & 0.034 & \textbf{0.861} & \textbf{0.265} & 0.254 & 0.028 \\
FakeShield~\citep{FakeShield} & 0.710 & 0.103 & 0.765 & 0.084 & 0.733 & 0.106 \\ \bottomrule
\end{tabular}
}
\vspace{-8pt}
\end{table}

We use Qwen-2.5-VL (7B- and 32B-Instruct variants) \citep{Qwen25VL} trained on 8x NVIDIA A100 GPUs. In the SFT stage, we employed a learning rate of $2 \times 10^{-5}$, whereas the subsequent GRPO stage utilized a learning rate of $10^{-5}$ with a group size of $G=4$. We used DeepSpeed ZeRO-3 during training.
For ablation, we evaluate single-turn variants of Qwen-2.5-VL-Instruct: the untrained base model (Base), a version using only explanations (E-), explanations with grounding but without refinement (E+G-), and our proposed LTE variants trained with SFT only, without GRPO.
For baselines, we compare our models against traditional classification methods, including Community Forensics~\citep{ComFor}, Antifake Prompt~\citep{AntifakePrompt}, DIRE~\citep{DIRE}, CNNSpot~\citep{CNNSpot} and NPR~\citep{NPR}.
For fair comparison, all baselines are retrained on TRACE's training split.
Due to their specific design, FakeShield~\citep{FakeShield} and LEGION~\citep{LEGION} cannot be directly trained or fine-tuned on TRACE because of incompatible dataset formats. Instead, we adopt the released pre-trained weights from FakeShield, and train LEGION on SynthScars~\citep{LEGION} to reproduce their results.

\subsection{Experimental Results}

\noindent\textbf{TRACE Results.}
In Table~\ref{tab:result-combined}, we report TRACE results for all VLM-based methods, including all LTE variants and the baselines LEGION~\citep{LEGION} and FakeShield~\citep{FakeShield}.
Accuracy is reported on the test split of the TRACE dataset.
The results show that our method's accuracy surpasses all existing detection methods, even with the smaller 7B variant.
Notably, compared to the original, untrained VLM baselines, our full pipeline with LTE reasoning yields an accuracy improvement of over 30\%. Furthermore, when compared to single-turn variants, the LTE mechanism consistently contributes an additional 3.6\% accuracy gain for the 7B and 5.8\% for the 32B model, validating our core hypothesis that progressive visual reasoning enhances detection, reaffirming the paradigm shift to reason and think \textit{with} images.

\noindent\textbf{OoD Results.} 
In addition to our dataset, we also evaluate other VLM-based methods on out-of-distribution (OoD) datasets with image and explanation pairs, including MMFR~\cite{FakeReasoning}, SynthScars~\cite{FakeShield} and FakeClue~\cite{fakeclue}.
Note that LEGION is trained on SynthScars, so its result on this dataset is not considered OoD.
The results in Table~\ref{tab:result-ood} show that our LTE models have an outstanding OoD performance, consistently surpassing most baseline models, and a high BLEU-2 performance indicates that explanations generated by LTE are semantically accurate and generalizable.

\begin{figure}[t]
\centering
\begin{minipage}[t]{0.48\textwidth}
    \centering
    \includegraphics[width=0.95\linewidth]{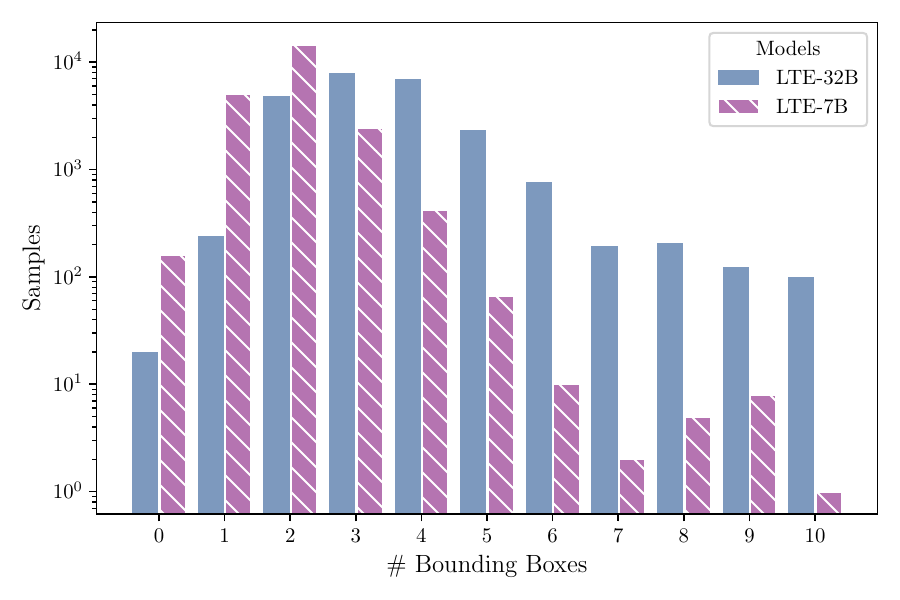}
    \caption{The number of bounding boxes in Query 1 for LTE-32B/7B models.}
    \label{fig:bbox-count}
\end{minipage}%
\hfill%
\begin{minipage}[t]{0.48\textwidth}
    \centering
    \includegraphics[width=0.95\linewidth]{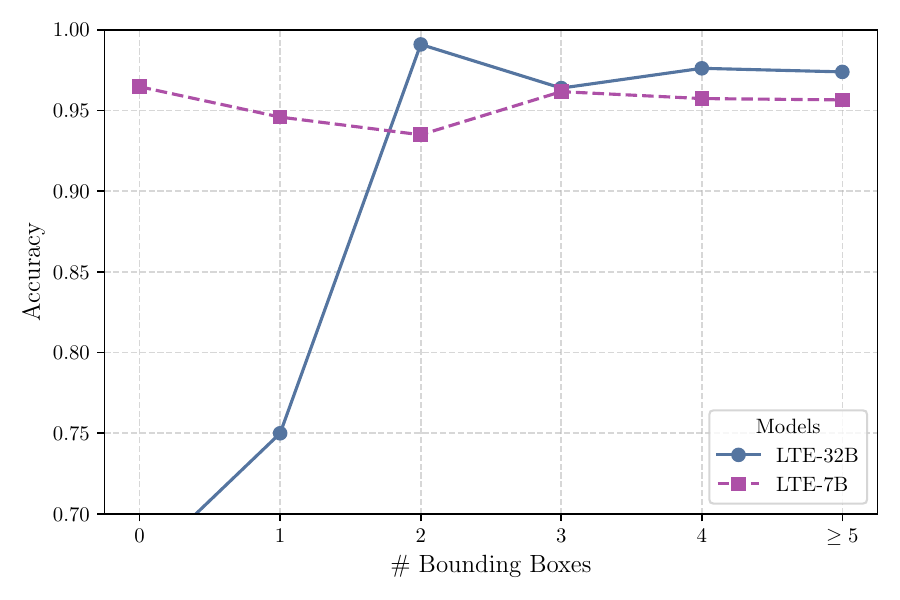}
    \caption{The relation of accuracy with regard to the number of detected bounding boxes.}
    \label{fig:bbox-count-acc}
    \vspace{-8pt}
\end{minipage}
\end{figure}

\noindent\textbf{Discussions on refinement corrections.} 
Figure~\ref{fig:bbox-count} demonstrates the correlation of accuracy and the number of LTE regions. We found that after training, the LTE-32B and 7B models exhibit different tendencies in Query 1.
In Query 1, the LTE-32B model outputs an average of 3.58 bounding boxes per image, while LTE-7B gives 1.95.
The 32B model tends to ask for two to four close-ups per image, summing up to 83.15\% among all input samples; meanwhile, the 7B model provides one or two bounding boxes for 86.34\% of the cases.
To further investigate the model performance with regard to the number of bounding boxes, Figure~\ref{fig:bbox-count-acc} shows that as the number of bounding boxes grows, the accuracy of LTE-7B slightly lowers, while LTE-32B has its accuracy spike at two bounding boxes, and performs consistently better than 7B on cases with more than two bounding boxes.
This phenomenon aligns with the fact that most training samples have an average of 3.24 bounding boxes.

\subsection{Ablation Studies}
\label{sec:ablation}
We conducted several ablation studies on the 32B model variant to validate our design choices. Table~\ref{tab:result-combined} shows the experimental results.
We report three metrics: the initial accuracy of the first turn (I-Acc.), the proportion of images forwarded to the second turn (C-Cases), and the corrected accuracy after refinement in the second turn (C-Acc.).

\noindent\textbf{Contribution of the LTE Mechanism.}
For LTE-7B and 32B models, 9.2\% and 10.9\% of the inputs will have a different verdict in Query 1 and Query 2 (i.e. $v_1\neq v_2$). Among these corrected cases, the accuracy is 91.5\% and 95.6\%, respectively, proving that models can refine their initial verdict with accurate grounding prior.

\noindent\textbf{Impact of BLEU Rewards.}
Removing linguistic quality rewards $R_B$ (``No BLEU Reward'') lowers the BLEU and ROUGE-L metrics as expected. In addition, we also notice a 4.3\% accuracy drop and a 6.3\% IoU drop.
This demonstrates that encouraging the model to generate coherent explanations improves not just text quality, but also its underlying forensic reasoning capabilities.

\noindent\textbf{Stage-Specific Rewards.}
We experimented with applying the classification accuracy reward ($\mathcal{R}_C$) during Query 1, in addition to Query 2.
The result in row ``Dual Verdict Reward'' shows a slight performance degradation.
The model became overly cautious, often failing to propose suspicious regions unless it was already highly confident in its initial verdict, thus undermining the purpose of the second-stage refinement.
Specifically, with this reward setting, the LTE process will no longer benefit the performance, and the IoU was also lower than the original reward setup.
This confirms that focusing Stage 1 rewards on localization is the optimal strategy in exploiting the VLM's intrinsic and learned capabilities in image forgery detection.
We also attempt to use the verdict as the only reward and remove the entire LTE process (``Verdict Only''). This results in slightly lower accuracy when compared to the E+G group and is more likely to overfit on the TRACE dataset.

\noindent\textbf{Random Cropping.}
Our two-stage inference pipeline is dependent on both the original image and cropped images (``Random Cropping'').
Therefore, we further ablate the bounding box selection stage and use random cropping instead.
While the model remains as LTE-32B, the bounding boxes given by the first query are no longer passed on to the second query.
Evaluation shows that when LTE operates on random regions, the accuracy can drop to 84.2\%, 13.0\% lower than the initial verdict.
BLEU-$n$ and ROUGE-L metrics also drop to a level similar to the base model.
This indicates that without intelligent region proposal, the LTE mechanism is ineffective and the explanation quality is low.

\noindent\textbf{Limiting Bounding Box Count.}
Since Figure~\ref{fig:bbox-count-acc} indicates that more bounding boxes may negatively impact model performance, we analyzed the impact of limiting the maximum number of proposed regions in Query 1 by selectively using the largest $n$ bounding boxes from the response.
As shown in Table \ref{tab:result-combined}, using the largest three or four bounding boxes can cause slightly reduced overall performance, with 1.8\% and 3.7\% drop in correction accuracy, respectively.

\subsection{Qualitative Analysis}

\begin{figure}
    \centering
    \includegraphics[width=\linewidth]{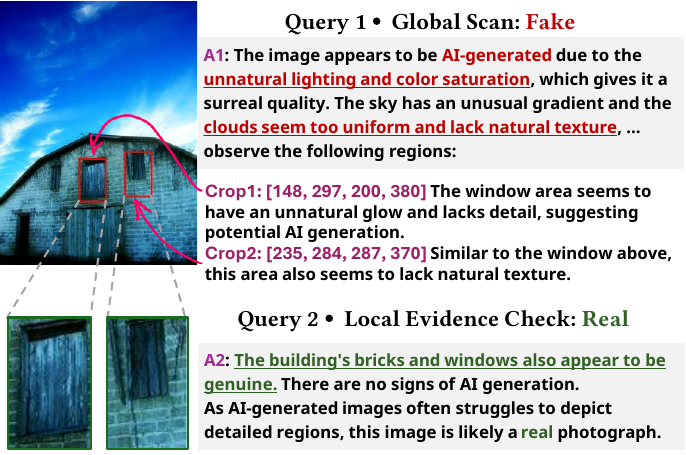}
    \caption{An example where LTE-32B corrects its initial mistake upon closer examination, finalizing with the correct conclusion that this image is real.}
    \label{fig:full-example}
    \vspace{-8pt}
\end{figure}

Figure~\ref{fig:samples} illustrates representative examples of successful outcomes, while Figure~\ref{fig:full-example} presents a specific ``corrected case''. 
In this instance, LTE-32B determines that the highly realistic and detailed rendering of window regions effectively refutes its initial assessment, thereby confirming that the image is likely a genuine photograph rather than an AI-generated picture.
Certain images with small text that are not clearly identifiable at first sight, minor anatomical issues (\textit{e.g.,} an extra finger), or texture problems are highly likely to be corrected during the refinement query.
We observe the following common patterns across successful predictions:

\noindent\textbf{Lighting inconsistencies} (12.4\%): Unnatural reflections, impossible shadows. \\
\noindent\textbf{Anatomical anomalies} (8.6\%): Distorted fingers, impossible joint positions. \\
\noindent\textbf{Texture artifacts \& blurry text} (8.1\%): Unnatural skin textures, clothing patterns. \\
\noindent\textbf{Perspective errors} (4.7\%): Impossible spatial relationships, such as overlapping object parts.

Failure cases often involve highly realistic AI-generated images, especially human or animal faces, where visual artifacts are subtle or imperceptible, even under magnification. Conversely, some real images with too complex textures or patterns are occasionally misidentified as synthetic.

\subsection{User Study}

To further validate the reasoning quality and ensure that our model is not overfitting on the lexical features of the dataset, we evaluate the results on the state-of-the-art VLMs, OpenAI GPT-5 and Gemini 2.5 Pro.
We ask the model to rate each explanation on a scale of one to five. We used the prompt:

\texttt{This image is \{label\}. The user will provide an explanation that verifies the authenticity of the image. Please rate its quality on a scale of 1 (worst) to 5 (best). Focus on the accuracy and completeness of the explanation. Your response must end with <score>X</score>.}

In addition to using LLM as judges, we also ask eight human experts that have experience in AI-generated art to rate the explanations of models, on the same scale of 1 to 5.
Each expert rated 300 responses, equally sourced from TRACE, LTE-7B and LTE-32B.
We randomly select the samples for each expert and ensure that no one will see an image twice. The image and explanation pair are provided via a dedicated web UI.
Each image and explanation pair is evaluated independently. Responses with wrong final verdicts are excluded.

\begin{table}[t]
\centering
\caption{The quality of explanations, rated on a 1–5 scale by LLM judges and human raters.}
\label{tab:user-study}
\resizebox{0.8\columnwidth}{!}{%
\begin{tabular}{@{}cccc@{}}
\toprule
\textbf{Judge}          & TRACE & LTE-7B & LTE-32B \\ \midrule
\textbf{OpenAI GPT-5}   & 3.57      & 3.15      & 3.53       \\
\textbf{Gemini 2.5 Pro} & 3.82      & 3.63      & 3.90       \\ \bottomrule
\textbf{Humans}         & 3.75      & 3.10      & 3.68       \\ \bottomrule
\end{tabular}
}
\vspace{-8pt}
\end{table}

Based on the results shown in Table~\ref{tab:user-study}, we can conclude that LTE-32B can explain with good quality, achieving scores that are highly comparable to the ground truth TRACE explanations when evaluated by both judge models as well as human experts.
This indicates that our model generates explanations at a quality nearly indistinguishable from the dataset's reference explanations, validating that our model has learned genuine forensic reasoning rather than simply overfitting to wordings and lingual patterns in the training data.

%% file: sec/5_conclusion.tex
\section{Conclusion}

In this paper, we propose Locate-Then-Examine, a two-stage VLM framework that detects AI-generated images by first localizing suspicious regions and then re-examining them to ground its final verdict in visual evidence. Experiments show that LTE achieves state-of-the-art accuracy on our benchmark and several out-of-distribution datasets, demonstrating that its refinement stage corrects initial errors and produces high-quality, interpretable explanations.

%% file: sec/X_suppl.tex
\clearpage
\setcounter{page}{1}
\maketitlesupplementary

\setcounter{section}{0}
\setcounter{figure}{0}
\setcounter{table}{0}
\renewcommand{\thesection}{\Alph{section}}

\section{More Details for the TRACE Dataset}

The TRACE dataset comprises 20,000 images (10,000 real and 10,000 AI-generated) annotated with forensic explanations and spatially grounded bounding boxes through an automated pipeline combining GPT-4o and Qwen-2.5-VL. This dataset addresses the critical need for grounding-aware training data in AI-generated image detection, particularly providing spatial localization capabilities alongside textual reasoning.

\subsection{Source of Images}

To ensure comprehensive coverage across different image categories and generation techniques, we sourced images from established datasets and state-of-the-art generation models.

\paragraph{Real Images.} The authentic images are sourced from two widely used computer vision datasets: ImageNet~\citep{ImageNet} and COCO~\citep{COCO}. These datasets provide diverse natural images spanning various object categories, ensuring broad coverage of real-world visual content. The selection from these datasets guarantees high-quality, authentic photographs that serve as reliable negative examples for training.

\paragraph{AI-Generated Images.} Half of the synthetic images are created using the OpenAI GPT-Image-1 model \citep{OpenAI4oGen}; the other half is generated by Gemini 2.5 Flash Image~\citep{NanoBanana}. Both models represent state-of-the-art text-to-image generation capabilities. This choice ensures that our dataset captures contemporary AI generation artifacts and challenges, providing relevant training examples for current detection scenarios.

\paragraph{Generation Prompts.}
All AI-generated images in TRACE are produced with a fixed photorealistic template: \texttt{"A realistic image of \{source text\}"}, where the source text is drawn from COCO captions or ImageNet class names.
COCO captions are typically full sentences containing one or more nouns and are descriptive enough to serve directly as generation input (e.g., \texttt{"A realistic image of a man riding a skateboard down a rail"}).
For ImageNet, we use the class name itself (e.g., \texttt{"A realistic image of a golden retriever"}).
No further prompt augmentation or variation is applied. This template is designed to steer the generators toward photorealistic outputs whose semantic content matches the real-image distribution, so that detection difficulty arises from visual artifacts rather than domain shift.

\subsection{Annotation Process}

We developed a completely automated pipeline that leverages the complementary strengths of different VLMs to generate comprehensive annotations without requiring extensive human labeling.

\begin{figure}[h]
    \centering
    \includegraphics[width=0.9\linewidth]{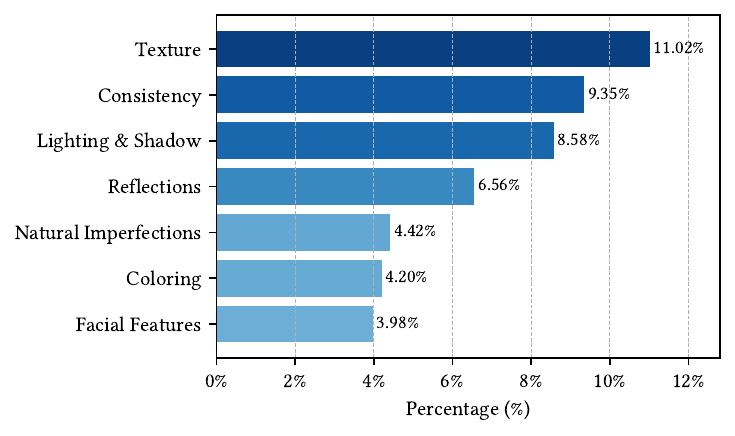}
    \caption{A statistical analysis of keywords in TRACE explanations.}
    \label{fig:artifact-attribution}
\end{figure}

\begin{figure}[t]
    \centering
    \includegraphics[width=0.95\linewidth]{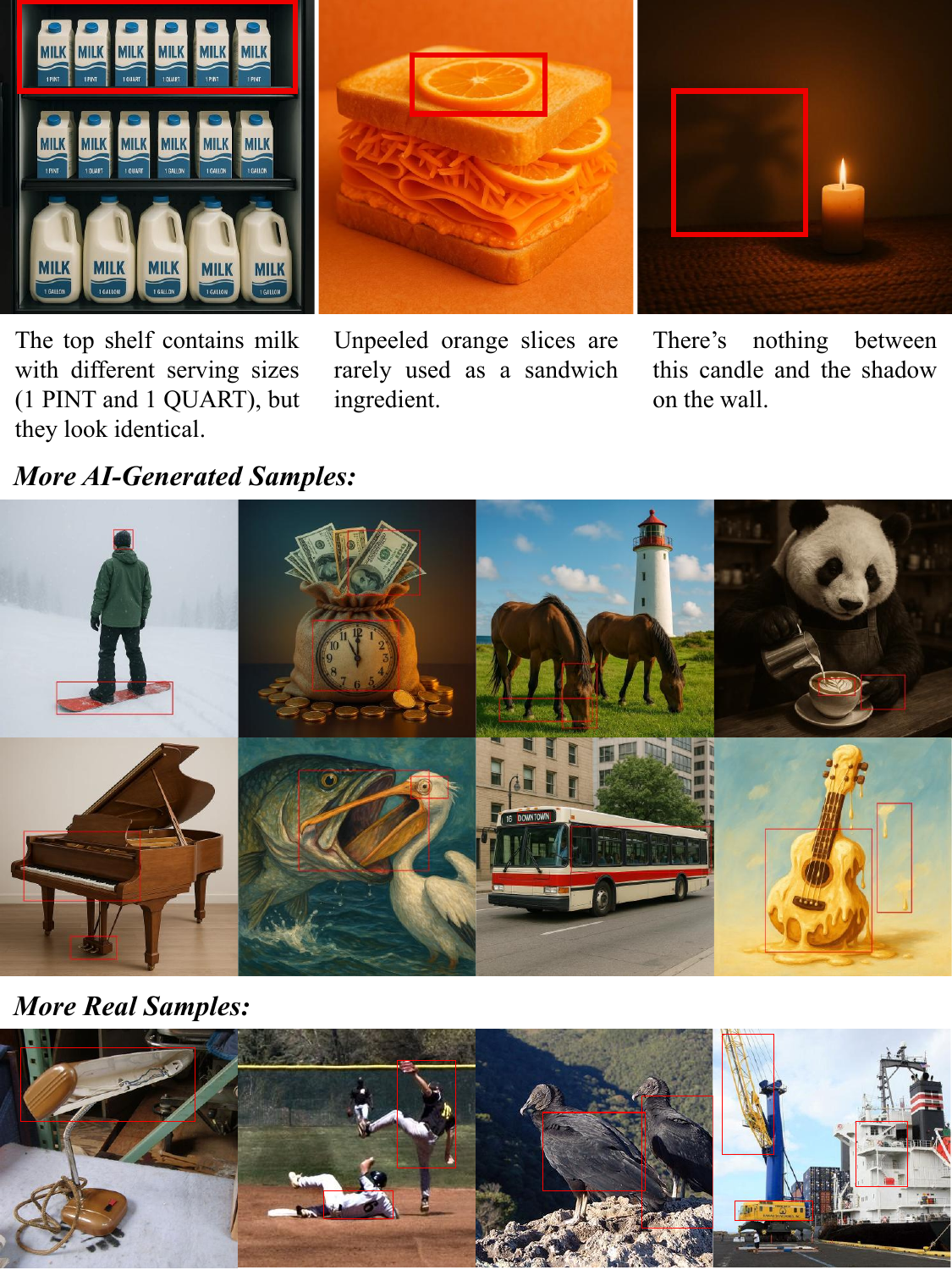}
    \caption{A collection of images from TRACE with rendered bounding boxes. The first row shows three AI-generated images with bounding boxes and corresponding explanations. The second row presents additional AI-generated samples, while the third row illustrates real images, all annotated with bounding boxes.}
    \label{fig:more-trace-examples}
\end{figure}

\begin{figure*}[ht]
    \centering
    \includegraphics[width=0.95\linewidth]{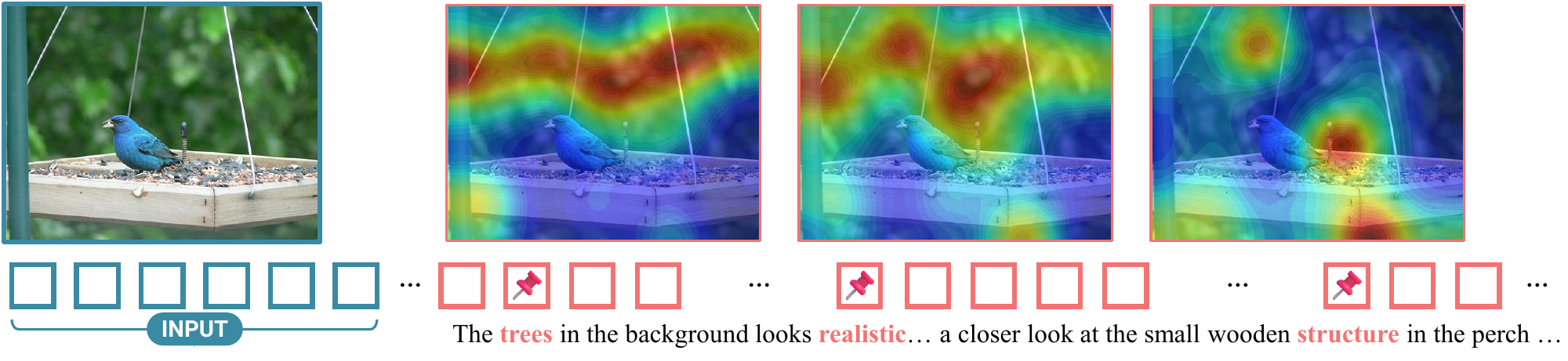}
    \caption{A visualization of the attention mechanisms of the VLM in Query 1.}
    \label{fig:gradcam}
\end{figure*}

\paragraph{Explanation Generation.} For images with known ground truth labels, GPT-4o generates detailed forensic explanations focusing on specific visual evidence that indicates whether an image is real or AI-generated. The prompts are designed to elicit detailed reasoning about depicted objects, spatial arrangements, perspective consistency, lighting patterns, and other forensic indicators. GPT-4o's strong reasoning capabilities and knowledge of image forensics make it well-suited for generating high-quality explanatory text that identifies key visual cues.

\paragraph{Spatial Grounding.} Qwen-2.5-VL extracts bounding boxes from the GPT-4o generated explanations, creating spatially grounded annotations in the form of $(I, y, E, B)$ tuples, where $I$ represents the image, $y\in\{\text{real},\text{generated}\}$ the ground-truth label, $E$ the explanation, and $B$ the bounding boxes. Image crops $C$ are deterministically derived from $(I, B)$ at training time. The Qwen-2.5-VL model demonstrates strong capabilities in extracting spatial regions based on textual descriptions, making it suitable for converting explanatory text into precise spatial coordinates.

\paragraph{Quality Control and Filtering.} During the spatial grounding phase, we observed that Qwen-2.5-VL occasionally generates bounding boxes encompassing over 50\% of the image area, often associated with global image characteristics such as over-saturation mentioned in the explanations. In some instances, the model reverts to object detection behavior when the primary flawed object occupies only a small portion of the image. To address these issues, we implement a filtering mechanism that leverages Qwen-2.5-VL to evaluate and remove bounding boxes that either fully encapsulate the primary object (indicating object detection regression) or cover an excessive portion of the image area. This filtration process ensures that the final annotations focus on specific forensic regions rather than global characteristics or entire objects.

\paragraph{Word Frequency Analysis.}
Figure~\ref{fig:artifact-attribution} displays the most frequently occurring keywords in the annotations. Texture and object consistency emerge as the primary concerns, followed by unnatural lighting, shadows, and reflections, indicating that these are the features most commonly leveraged by the model for detecting synthetic content.

\subsection{More Samples from TRACE}

Figure~\ref{fig:more-trace-examples} presents additional examples from the TRACE dataset, demonstrating the diversity of forensic indicators captured by our automated annotation pipeline.
The first row shows three AI-generated images with a brief summary of the explanation.
We can see that the explanations cover both fine-grained and general semantic reasoning of why this image should be considered real or AI-generated.
TRACE features both real and AI-generated images. Four real image samples are provided at the bottom row.
This dataset covers a wide range of reasons, fostering explainability for fine-tuned models, and also demonstrating the complexity and variability inherent in synthetic imagery.  

\subsection{Ethical Considerations}

All AI-generated images in the dataset are created specifically for this research and do not depict real individuals.
The real images sourced from ImageNet and COCO are used in accordance with their respective licensing terms and ethical guidelines.

\subsection{Known Limitations}

\paragraph{Automated Annotation Bias.} While our automated pipeline reduces human annotation costs, it may inherit biases from the underlying VLMs used for annotation. The quality of explanations and spatial grounding depends on the capabilities and training data of GPT-4o and Qwen-2.5-VL, potentially limiting the coverage of subtle or novel forensic indicators.
\paragraph{Language Limitation.} All explanations are generated in English, limiting the applicability of the dataset for multilingual forensic applications. Translation of the nuanced forensic explanations to other languages would require careful validation to maintain technical accuracy.

\section{VLM Attention Visualization}

In our multi-stage reasoning pipeline, we aim to enable the model to actively ``look'' for suspicious or diagnostically relevant regions within images, thereby facilitating deeper, more focused analysis in the second step.
A natural question arises: can the VLM truly identify image regions that are semantically aligned with the generated text and relevant to the real/fake detection task?

To verify whether our VLM, Qwen-2.5-VL-32B-Instruct, is indeed attending to specific, meaningful patches of the input image, rather than relying solely on global context or textual priors, we conducted a visualization study using gradient-based attention mapping on a representative sample from Query 1.
Specifically, we generated LLaVA-CAM~\citep{LLaVA-CAM} heatmaps to highlight the regions of the image that most strongly influence the model’s output predictions.
As shown in Figure~\ref{fig:gradcam}, there is a clear and compelling correspondence between the highlighted areas in the heatmap and the content of the model’s generated textual explanation, confirming that the model can localize and focus on relevant regions, which lays a solid foundation for the LTE process.
Moreover, we observe that attention often centers around keywords (\textit{e.g.}, ``realistic'') in the textual explanation, reinforcing the connection between visual grounding and real/fake decision-critical semantics. This confirms that the model can meaningfully propose LTE regions that support reliable second-stage analysis.

\section{More Experimental Details}

Since real and AI-generated images are not of the same resolution or aspect ratio, we performed center-cropping and resizing to ensure all input images have a resolution of $512\times 512$ during training.

We use ms-swift to fine-tune VLMs.
The batch size is set to 1. 
During the GRPO stage, the number of generations is set to 2.
For LTE-32B, the full training pipeline took 42.6 hours on 8x NVIDIA A100 GPUs. We found that at least 600 GB of VRAM is required to perform GRPO.
For LTE-7B, the training took 35.3 hours on 4x NVIDIA A100 GPUs.
The training process is generally stable. A few loss spikes are observed during the first 1,000 steps of training, but the model quickly converges after that and recovers from the spike.

\paragraph{Details of Baseline Methods}
A range of methodologies has been proposed for detecting synthetic content, each grounded in distinct theoretical assumptions and detection paradigms.

\textit{CNNSpot}~\citep{CNNSpot} hypothesizes that CNN-based generative models leave consistent, detectable artifacts and achieve cross-generator generalization through data augmentation.
We trained CNNSpot from scratch on the training set of TRACE. The training settings are the same as described in the original work.

\textit{Community Forensics}~\citep{ComFor} adopts a data-centric approach, positing that detection performance scales with the diversity and quantity of training generators, and introduces a large-scale dataset comprising thousands of generators to train robust classifiers.

\textit{DIRE}~\citep{DIRE} takes a process-centric perspective, exploiting the asymmetric reconstruction behavior of diffusion models: real and generated images exhibit differing error patterns when reverse-denoised, forming a discriminative signal known as the DIRE map.

\textit{Antifake Prompt}~\citep{AntifakePrompt} leverages VLMs and reformulates detection as a visual question-answering task, employing parameter-efficient soft prompt tuning on a frozen VLM to enable generalization.

\textit{NPR}~\citep{NPR} utilizes neighboring pixel relationships to identify AI-generated images with good accuracy and generalizability, as CNN-based generative methods exhibit patterns in neighboring pixels.

Collectively, these methods represent diverse strategies from artifact analysis to semantic reasoning, advancing the state of synthetic content detection.
During evaluation, all models are trained on the training set of TRACE with the same setup as the original work.

\begin{table*}[t]
\centering
\caption{Performance on TRACE with degradation, including JPEG compression artifacts, random cropping and image down-sampling.}
\label{tab:robustness}
\resizebox{0.85\linewidth}{!}{%
\begin{tabular}{@{}cccccccccc@{}}
\toprule
Degradation & Metric & \textbf{LTE} & FakeShield & LEGION & ComFor. & AfPr. & DIRE & CNNSpot & NPR \\ \midrule
\multirow{2}{*}{\begin{tabular}[c]{@{}c@{}}JPEG Compression\\ (80\% Quality)\end{tabular}} & Acc. & \textbf{0.970} & 0.781 & 0.518 & 0.832 & 0.873 & 0.913 & 0.849 & 0.869 \\
 & IoU & \textbf{0.355} & 0.089 & 0.067 & - & - & - & - & - \\ \midrule
\multirow{2}{*}{\begin{tabular}[c]{@{}c@{}}JPEG Compression\\ (30\% Quality)\end{tabular}} & Acc. & \textbf{0.964} & 0.768 & 0.505 & 0.791 & 0.852 & 0.896 & 0.837 & 0.835 \\
 & IoU & \textbf{0.347} & 0.086 & 0.066 & - & - & - & - & - \\ \midrule
\multirow{2}{*}{Random Cropping} & Acc. & \textbf{0.965} & 0.756 & 0.513 & 0.835 & 0.877 & 0.909 & 0.848 & 0.866 \\
 & IoU & \textbf{0.306} & 0.061 & 0.063 & - & - & - & - & - \\ \midrule
\multirow{2}{*}{\begin{tabular}[c]{@{}c@{}}Downsampling\\ (0.5x)\end{tabular}} & Acc. & \textbf{0.969} & 0.759 & 0.514 & 0.890 & 0.886 & 0.912 & 0.851 & 0.874 \\
 & IoU & \textbf{0.346} & 0.075 & 0.070 & - & - & - & - & - \\ \bottomrule
\end{tabular}%
}
\end{table*}

\begin{figure}[t]
\centering
\begin{minipage}[t]{0.48\textwidth}
    \centering
    \includegraphics[width=\linewidth]{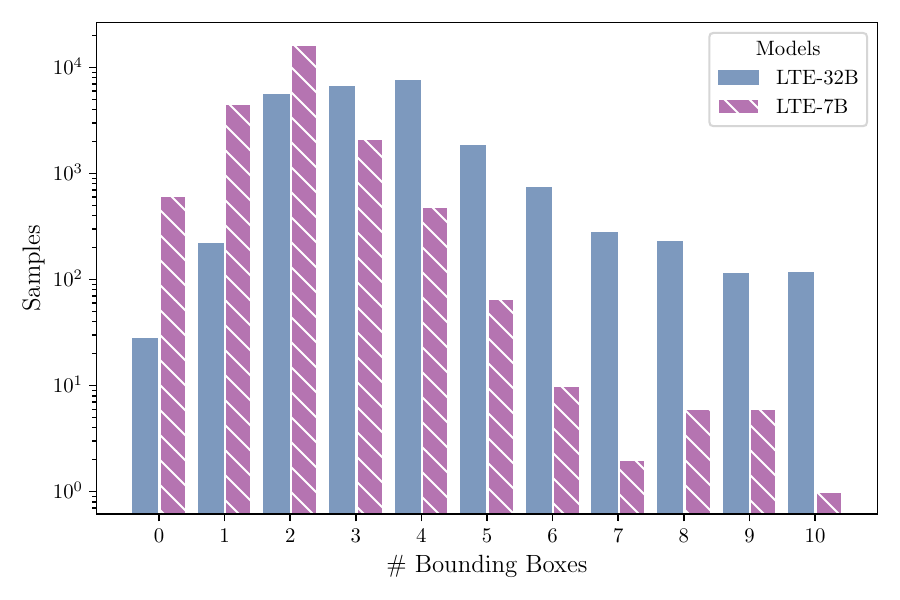}
    \caption{Number of samples grouped by the bounding boxes on OoD datasets.}
    \label{fig:bars}
\end{minipage}%
\hfill%
\begin{minipage}[t]{0.48\textwidth}
    \centering
    \includegraphics[width=\linewidth]{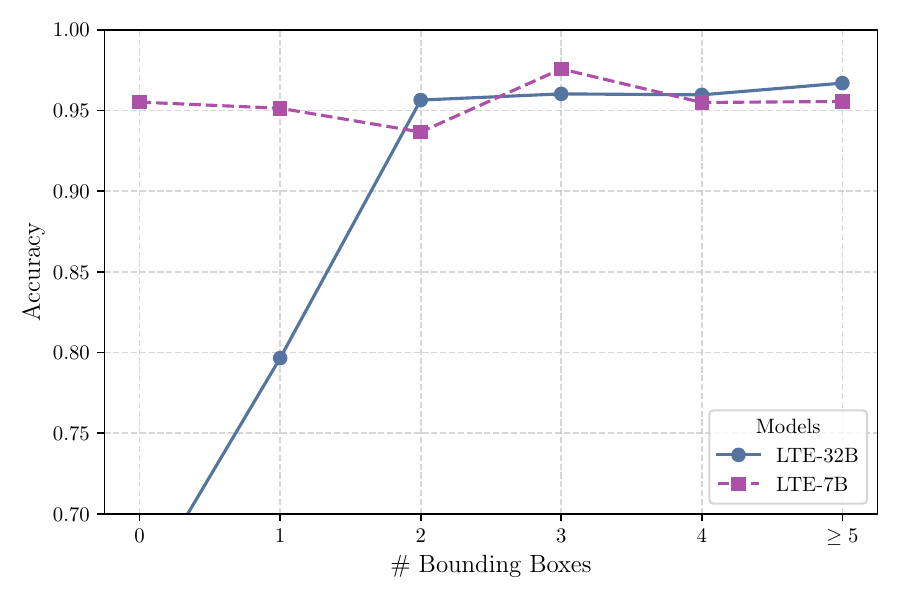}
    \caption{The relation of accuracy with regard to the number of detected bounding boxes on OoD datasets.}
    \label{fig:curves}
\end{minipage}
\end{figure}
\section{Analysis of Bounding Boxes on OoD Datasets} 

We collected the responses from LTE models when evaluating on OoD datasets, GenImage~\citep{GenImage}, MMFR-Dataset~\citep{FakeReasoning} and SynthScars~\citep{LEGION}.
On these OoD datasets, Figure~\ref{fig:bars} and~\ref{fig:curves} display the relation of bounding boxes with regard to model performance, and the number of detected bounding boxes for each LTE model variant (7B and 32B).
The trend of Figure~\ref{fig:bars} highly resembles  \textit{Figure 5} in the main paper, while Figure~\ref{fig:curves} is slightly different than \textit{Figure 6}, where 32B performs better than 7B for cases with more regions selected.
This proves that knowledge from the TRACE dataset can be adapted beyond the dataset, and 32B generalizes better than 7B on OoD datasets.

\section{Robustness Against Degradations}

We evaluate the robustness on TRACE dataset under four common degradations: JPEG compression at 80\% and 30\% quality, random cropping, and 0.5$\times$ downsampling (Table~\ref{tab:robustness}). All methods exhibit performance drops relative to clean images, with the extent varying across perturbations and models.

LTE attains the highest accuracy in every setting and the best IoU among methods that produce localization, with modest declines across degradations (IoU: 0.355 at JPEG~80\%, 0.347 at JPEG~30\%, 0.346 under downsampling, and 0.306 with random cropping). Random cropping is most detrimental to localization quality, while heavy JPEG compression tends to reduce classification accuracy the most for several baselines. Among the baselines, DIRE consistently produces the best results, followed by AntifakePrompt and CommunityForensics, whereas all VLM-based methods show good robustness against JPEG compression.
Downsampling by 50\% affects the performance across the models nonuniformly.

These results indicate that, despite relatively strong detection accuracy, most current detectors remain sensitive to common real-world degradations, while VLM-based methods have a lower rate, suggesting opportunities for improvement via degradation-aware training, stronger invariances, and reduced reliance on dataset-specific biases.

\section{Computational Efficiency}

\begin{table}[t]
\centering
\caption{Mean and standard deviation of inference time per image.}
\label{tab:result-efficiency}
\resizebox{4.7cm}{!}{%
\begin{tabular}{@{}cc@{}}
\toprule
Method              & Seconds / Image       \\ \midrule
\textbf{LTE-32B} & \textbf{24.0$\pm$3.0} \\
E-32B (one-turn)    & 10.9$\pm$2.2          \\
\textbf{LTE-7B}  & \textbf{8.94$\pm$1.21} \\
Base-7B (one-turn)  & 4.38$\pm$0.71          \\ \midrule
LEGION              & 11.3$\pm$2.61        \\
FakeShield          & 60.4$\pm$5.33        \\
NPR                 & 0.105$\pm$0.02        \\
AntifakePrompt      & 0.182$\pm$0.05        \\ \bottomrule
\end{tabular}
}
\end{table}

Table \ref{tab:result-efficiency} lists the end-to-end inference time for our trained VLMs.
We use vllm~\citep{VLLM} to accelerate the inference process.
The deployment of LTE-32B takes 4x NVIDIA A100-40G GPUs connected with PCI-E.
LTE-7B, however, is deployed on one NVIDIA A100-40G GPU.
While our two-stage approach increases inference time from traditional classification methods, this overhead is justified by accuracy improvements and interpretability gains. For high-stakes forensic applications, the trade-off favors thoroughness over speed.

\section{Additional Ablation Studies}

\paragraph{LLM-Based Explanation Reward.}
To verify that the BLEU-based reward $\mathcal{R}_{\text{BLEU}}$ does not cause lexical overfitting, we train an alternative model, LTE$^+$-7B, that replaces $\mathcal{R}_{\text{BLEU}}$ with an LLM-based semantic reward $\mathcal{R}_{\text{LLM}}$.
Specifically, GPT-5 evaluates each generated explanation against the ground truth on a 0--1 scale, scoring factual consistency rather than surface-level $n$-gram overlap.

\begin{table}[t]
\centering
\caption{BLEU-based vs.\ LLM-based explanation rewards on the TRACE test set (7B models).}
\label{tab:llm-reward}
\resizebox{0.6\columnwidth}{!}{%
\begin{tabular}{@{}lccc@{}}
\toprule
Setting & Acc. & BLEU-2 & IoU \\ \midrule
SFT-7B & 0.719 & 0.042 & 0.094 \\
LTE-7B & \textbf{0.942} & \textbf{0.209} & \textbf{0.316} \\
LTE$^+$-7B & 0.937 & 0.191 & 0.310 \\ \bottomrule
\end{tabular}%
}
\end{table}

As shown in Table~\ref{tab:llm-reward}, LTE$^+$-7B reaches 93.7\% accuracy and 31.0\% IoU, closely matching LTE-7B trained with $\mathcal{R}_{\text{BLEU}}$.
Both post-RL models substantially outperform SFT-7B across all metrics.
The small gap between the two reward formulations suggests that $\mathcal{R}_{\text{BLEU}}$ serves as a reasonable proxy for semantic quality in this task and does not bias the model toward shallow lexical mimicry.

\paragraph{Data Scaling.}
We examine how training set size affects the two-stage framework by training 7B models on randomly sampled TRACE subsets of 5K and 10K images (2 epochs SFT + 2 epochs GRPO each). One-turn (1T) variants that skip the examine stage are included for comparison.

\begin{table}[t]
\centering
\caption{Effect of training data scale on detection and grounding (7B models, TRACE test set).}
\label{tab:data-scaling}
\resizebox{0.65\columnwidth}{!}{%
\begin{tabular}{@{}lccc@{}}
\toprule
Setting & Acc. & BLEU-2 & IoU \\ \midrule
LTE-7B-5K & 0.886 & 0.154 & 0.258 \\
LTE-7B-5K (1T) & 0.857 & 0.155 & 0.235 \\
LTE-7B-10K & 0.925 & 0.188 & 0.269 \\
LTE-7B-10K (1T) & 0.883 & 0.183 & 0.252 \\
LTE-7B & \textbf{0.942} & \textbf{0.209} & \textbf{0.316} \\ \bottomrule
\end{tabular}%
}
\end{table}

Table~\ref{tab:data-scaling} shows a clear upward trend as training data grows from 5K to the full 20K split, with gains across accuracy, explanation quality, and grounding precision. At every scale, the two-turn LTE pipeline outperforms its one-turn counterpart, confirming that the locate-then-examine design is effective independent of dataset size and not an artifact of overfitting to the full training set.

\section{Limitations \& Future Works}

The generalization capability of our method has not yet been thoroughly evaluated. In future work, we aim to conduct more comprehensive assessments using a broader range of datasets and image sources to better understand the model’s detection accuracy across diverse scenarios. 

While our current approach equips the model with a cropping tool to facilitate image-based reasoning, this represents only a preliminary step toward enabling true visual thinking. Despite its effectiveness, there remains significant potential for further exploration in this direction, particularly in developing more sophisticated interactive mechanisms that empower the model to dynamically analyze and reason over visual content.